\documentclass[11pt]{article}
\usepackage[margin=1.1in]{geometry}
\usepackage{graphicx}
\usepackage{booktabs}
\usepackage{amsmath}
\usepackage{lmodern}
\usepackage{microtype}
\usepackage{caption}
\usepackage[hidelinks,backref=page]{hyperref}
\usepackage[capitalise,noabbrev]{cleveref}
\usepackage{url}
\title{\bf Jev Matches 7B Language Models for Speech-Neuroprosthesis Rescoring}
\author{Gabriele Cin\`a\\\small\texttt{c.gabriele.info@gmail.com}}
\date{}

\begin{document}
\maketitle

\begin{abstract}
\noindent
A speech neuroprosthesis decodes attempted speech from brain activity and ends by
rescoring the decoder's candidate sentences with a language model of several billion
parameters, the only component that needs a GPU. Replacing that model with a cheaper one
is hard: general language models asked to pick one sentence from a list answer from where
a label sits in the list rather than from the sentence itself. We pose rescoring as a
single typed decision, one call that returns a probability for every candidate, served by
Jev, a hosted model trained for calibrated decisions, and combine it with the decoder's
own score. On 978 held-out sentences from a participant with ALS, where the published
decoder alone reaches 8.1\% word error, Jev reaches 7.5\% against 7.8\% for both OPT-6.7b
and Qwen2.5-7B; with the decoder's weight re-tuned, 6.9\% against 7.2\% and 7.4\%. Jev is
ahead in all four comparisons and at most 0.2 points behind at the 95\% bound. It costs
\$0.07 per thousand sentences and needs no GPU; a dedicated GPU running a 7B model is
cheaper per sentence only above 43\% utilisation, far beyond what one user generates.
End-to-end latency over the internet is 262\,ms, of which 62\,ms is spent at the provider,
the same order as a 7B model on a local GPU (27\,ms) but not faster.
\end{abstract}

\section{Introduction}

A speech neuroprosthesis turns intracortical activity into text in three stages. A
recurrent network trained with connectionist temporal classification~\cite{graves2006ctc}
emits phoneme probabilities, a lexicon- and $n$-gram-constrained decoder turns them into a
ranked list of candidate sentences, and a language model rescores that list before the
winner is spoken~\cite{willett2023,card2024}. The Brain-to-Text benchmark's reference
implementation rescores with OPT-6.7b~\cite{zhang2022opt,b2tbench}. This last stage is the
only one that needs a multi-billion-parameter model, so it sets the accelerator, the power
budget and the cost of the system.

\begin{figure}[t]
\centering
\includegraphics[width=0.72\linewidth]{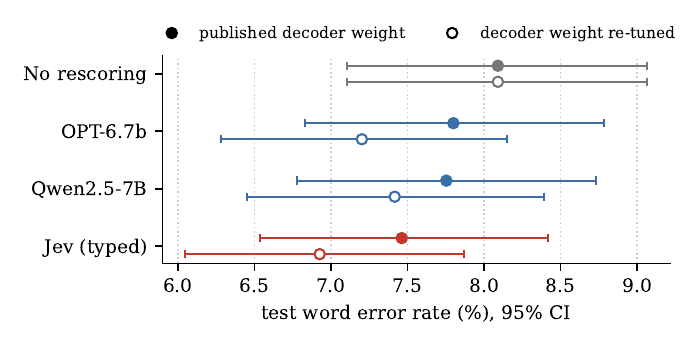}
\caption{Test word error with 95\% bootstrap intervals, 978 sentences, identical
candidate lists. Marginal intervals overlap because they share sentence-level noise; the
paired tests in \cref{tab:acc,tab:pair} hold the sentences fixed and are the sensitive
comparison.}
\label{fig:wer}
\end{figure}

The rescorer's job is selection: the answer is one of $N$ sentences already written down.
A model that only selects should suffice, and one call over the whole list is the natural
shape for it. General language models handle that shape badly. Asked to choose from a
list, they favour positions over contents~\cite{tang2023psc,bito2026position}, and in this
task open models posed this way place most of their answers on the decoder's own top
candidate and almost never pick the truth when they deviate from it (\cref{sec:collapse}).

A typed-decision model is trained for this output type. Given a state and a question with
a fixed set of answers, it returns a probability for each answer in one pass and cannot
produce anything outside the set. Jev~\cite{typesafe2026blog} is such a model, trained
with reinforcement learning for calibrated decisions and served through an API. Early
third-party evaluations report lower latency and cost than API-served language models on
service orchestration~\cite{li2026edge} and text annotation~\cite{ibrahim2026annotation},
but none tests it on neural decoding or against locally hosted scorers.

We test Jev as the rescoring stage of the published pipeline for participant T15. Every
arm scores byte-identical cached candidate lists, so the rescorer is the only variable. We
compare against the two 7B models the stage would otherwise use, under the published
decoder setting and with the decoder's weight re-tuned, and measure cost and latency.

\paragraph{Contributions.}
\begin{itemize}\itemsep2pt
\item On real intracortical recordings, a typed-decision model matches two 7B rescorers:
  it is ahead of OPT-6.7b and Qwen2.5-7B under both tuning protocols, and the worst-case
  95\% bound puts it 0.2 points behind (\cref{sec:acc}).
\item Posing the same typed decision to open general models fails: their deviations from
  the decoder's top candidate select the truth 0.2--2.4\% of the time against 26.6\% for
  Jev (\cref{sec:collapse}).
\item Jev costs \$0.07 per thousand sentences with no local accelerator, and its API
  latency is dominated by transport rather than by the model (\cref{sec:cost}).
\end{itemize}
Cached lists, per-arm scores, latency logs and code are
released.\footnote{\url{https://github.com/gabrycina/how-much-language-model}}

\section{Related work}
\label{sec:related}

\paragraph{Rescoring in speech neuroprostheses.} Willett et al.~\cite{willett2023} rescore
with a large language model and report word error falling from 23.8\% to 17.4\%; Card et
al.~\cite{card2024} use the same stage in a system that reaches single-digit error. Both
use multi-billion-parameter models. We keep their pipeline and replace only the rescorer.

\paragraph{Rescoring in speech recognition.} Automatic speech recognition interpolates the
first-pass score with a language model's likelihood~\cite{udagawa2022} or trains the
rescorer discriminatively~\cite{rescorebert,shivakumar2023disc,lob2023}. Those methods
apply here in principle, but they need training lists at a scale one participant's
419-sentence development split cannot supply; in our experiments, minimum-word-error
fine-tuning of GPT-2 small on 300 lists did not beat the decoder alone. Every arm here is
therefore untuned, Jev included.

\paragraph{Listwise selection with language models.} Listwise reranking asks a model to
order or pick from a list in one pass. Its known failure is position bias, which
permutation self-consistency~\cite{tang2023psc} and order-invariant
architectures~\cite{bito2026invarirank} address. On an earlier decoder we applied
permutation self-consistency to the open typed arms; neither beat the decoder's own top
candidate.

\paragraph{Typed-decision models.} Li et al.~\cite{li2026edge} replace a language model
with Jev for edge service orchestration and report 16--27\% lower median latency and about
70\% lower cost per correct completion than an API-served language model. Ibrahim and
Zaki~\cite{ibrahim2026annotation} evaluate decision models for text annotation and find
Jev ahead of an open 0.6B model trained with the same objective. Both compare against
generation over an API. Our baselines are scorers on a local GPU, which is the harder
comparison for latency.

\section{Background}
\label{sec:background}

For each attempted sentence the decoder emits a list $H = \{h_1, \dots, h_N\}$ ranked by
\begin{equation}
  \ell(h) = \lambda\, a(h) + g(h),
\end{equation}
where $a(h)$ is the acoustic score from the phoneme network, $g(h)$ the $n$-gram score and
$\lambda$ the acoustic weight, 0.3 in the published pipeline.

A rescorer assigns $r(h)$ to every candidate, and the system speaks
\begin{equation}
  \hat h = \arg\max_{h \in H}\; (1-\alpha)\, z_H[\ell](h) + \alpha\, z_H[r](h),
\end{equation}
where $z_H$ standardises a score to zero mean and unit variance across the list, so that
rescorers on different scales share one weight $\alpha \in [0,1]$.

A per-candidate rescorer sets $r(h)$ to the length-normalised log-likelihood of $h$ under
a language model, one sequence per candidate. A typed rescorer receives the whole list
once, with candidate $h_i$ under label $i$, and returns a distribution $p$ over labels; we
set $r(h_i) = \log \max(p_i, 10^{-6})$.

\section{Experimental setup}
\label{sec:setup}

\paragraph{Data and candidate lists.} We use the T15 recordings released with the
Brain-to-Text benchmark~\cite{b2tbench,card2024}: 256 electrodes over speech motor cortex,
decoded by the authors' pretrained recurrent network, unmodified. The published Kaldi
decoder with an OpenWebText 3-gram produces 100-best lists for the 1,397 sentences of the
benchmark's validation split (oracle word error 2.5\%, truth in the list 89\%). A seeded
split gives 419 development and 978 test sentences. Lists are cached, so every arm scores
identical candidates.

\paragraph{Arms.} OPT-6.7b~\cite{zhang2022opt} and Qwen2.5-7B score every candidate on one
NVIDIA L40S in bfloat16, each candidate as a standalone sentence, in batches of 32.
Jev (\texttt{jev-latest}, September 2026) receives the list once through its API
(\cref{app:prompt}). As controls, Laya, an open 421M typed-decision model, receives the
identical request, and OPT-6.7b and Qwen2.5-7B receive the same task and list as a text
prompt, read from their probability for each label as the next token.

\paragraph{Tuning.} Under the \emph{published} protocol $\lambda = 0.3$ and $\alpha$ is
tuned per arm over 21 values in $[0,1]$. Under the \emph{re-tuned} protocol $\lambda$ is
searched jointly over 17 values in $[0,5]$. Both protocols select on the development split
only and score the test split once. Without a rescorer the development split selects
$\lambda = 0.3$ in both protocols.

\paragraph{Statistics.} Word error is corpus-level Levenshtein distance over words
(jiwer 4.0). Intervals are percentile bootstrap over sentences (2,000 resamples);
comparisons are one-sided paired bootstraps (5,000) on the test split, each arm at its own
tuned setting, with Holm correction over the three arm-versus-baseline tests within each
protocol. We say Jev \emph{matches} a 7B model when the 95\% interval of their paired
difference excludes a Jev deficit of more than 0.2 points. We fixed this bound after
seeing the results (\cref{sec:limits}).

\paragraph{Cost and latency.} Jev's list price is \$0.042 per million input tokens with
no charge for output; we read token counts from the API's usage field on 140 random test
lists. The GPU is an on-demand L40S at \$1.82 per hour. All arms were timed per list
during scoring, from one data centre in Finland. We also timed Jev from London on 80 test
lists, each paired with a two-option request, and recorded the provider's
\texttt{x-envoy-upstream-service-time} header, the time its proxy waits for the model
service.

\section{Results}
\label{sec:results}

\subsection{Accuracy}
\label{sec:acc}

\begin{table}[t]
\centering\small
\begin{tabular}{lrrrrrr}
\toprule
& \multicolumn{3}{c}{published ($\lambda = 0.3$)} & \multicolumn{3}{c}{re-tuned $\lambda$} \\
\cmidrule(lr){2-4}\cmidrule(lr){5-7}
Arm & WER & $\Delta$ & $p_{\text{Holm}}$ & $(\lambda, \alpha)$ & WER & $\Delta$ \\
\midrule
No rescoring & 8.09 & --- & --- & (0.3, ---) & 8.09 & --- \\
OPT-6.7b & 7.80 & $-0.29$ & 0.070 & (0.40, 0.45) & 7.20 & $-0.89$ \\
Qwen2.5-7B & 7.75 & $-0.34$ & 0.070 & (0.45, 0.45) & 7.42 & $-0.67$ \\
Jev (typed) & \textbf{7.46} & $-0.63$ & 0.018 & (2.0, 0.75) & \textbf{6.93} & $-1.17$ \\
\bottomrule
\end{tabular}
\caption{Test word error (\%) and paired difference from no rescoring. Under the
re-tuned protocol all three arms separate from the baseline (Holm-adjusted $p \le 0.002$).
Published-protocol $\alpha$: OPT 0.30, Qwen 0.45, Jev 0.90.}
\label{tab:acc}
\end{table}

\begin{table}[t]
\centering\small
\begin{tabular}{llrlr}
\toprule
Protocol & Comparison & $\Delta$WER & 95\% CI & $p$ \\
\midrule
published & Jev $-$ OPT-6.7b & $-0.34$ & $[-0.80, +0.12]$ & 0.082 \\
published & Jev $-$ Qwen2.5-7B & $-0.29$ & $[-0.78, +0.20]$ & 0.129 \\
re-tuned & Jev $-$ OPT-6.7b & $-0.28$ & $[-0.72, +0.15]$ & 0.112 \\
re-tuned & Jev $-$ Qwen2.5-7B & $-0.49$ & $[-0.98, -0.03]$ & 0.019 \\
\bottomrule
\end{tabular}
\caption{Jev against each 7B rescorer, paired on the test split. Negative favours Jev. The
largest upper bound, $+0.20$, is the matching margin.}
\label{tab:pair}
\end{table}

Jev is the most accurate arm under both protocols (\cref{fig:wer}, \cref{tab:acc}). At
the published setting it reaches 7.46\% against 7.80\% for OPT-6.7b and 7.75\% for
Qwen2.5-7B, and it is the only arm that separates from no rescoring after correction.
With $\lambda$ re-tuned every arm improves and all three separate: Jev reaches 6.93\%,
OPT-6.7b 7.20\% and Qwen2.5-7B 7.42\%.

Paired against each 7B model, Jev is ahead in all four comparisons by 0.28--0.49 points
(\cref{tab:pair}). Only the re-tuned comparison with Qwen2.5-7B reaches $p < 0.05$, and it
does not survive correction over the four tests (adjusted $p = 0.076$). The data therefore
support equivalence rather than superiority: across both models and both protocols the
95\% intervals exclude a Jev deficit larger than 0.20 points.

The tuned weights show how Jev is used. Its $\alpha$ is 0.90 at the published setting
and 0.75 after re-tuning, against 0.30--0.45 for the 7B models, so the combination leans
mostly on Jev. Re-tuning also moves Jev's $\lambda$ to 2.0, weighting the acoustic score
far above the $n$-gram. A plausible reading is that Jev takes over the $n$-gram's role
of judging which candidate reads as English, leaving the decoder to supply acoustic
evidence.

\subsection{The typed decision needs a trained model}
\label{sec:collapse}

\begin{table}[t]
\centering\small
\begin{tabular}{lrrrrr}
\toprule
Typed arm & Labels used & Share on A & Truth picked & Truth picked ($\neq$ A) & WER \\
\midrule
Jev & 57 / 100 & 72\% & 67.9\% & 26.6\% & 7.46 \\
Laya (421M) & 44 / 100 & 69\% & 53.7\% & 2.3\% & 8.09 \\
Qwen2.5-7B & 31 / 100 & 28\% & 29.3\% & 2.4\% & 8.11 \\
OPT-6.7b & 13 / 100 & 35\% & 27.3\% & 0.2\% & 8.09 \\
\midrule
\emph{Decoder's top candidate} & --- & --- & \emph{68.0\%} & --- & \emph{8.09} \\
\bottomrule
\end{tabular}
\caption{The same typed task given to four models, all 1,397 lists. Label A is the
decoder's top candidate. ``Truth picked ($\neq$ A)'' conditions on the model choosing any
other label. WER is on the test split at the published setting, $\alpha$ tuned; the
three open arms stay at the baseline.}
\label{tab:collapse}
\end{table}

The gain comes from the model, not from the typed format. Given the same task, the
three open models land on the baseline (\cref{tab:collapse}); OPT-6.7b and Laya tune to
$\alpha = 0$, so the combination discards them. All four typed arms place many answers on
label A, the decoder's own top candidate. What separates them is what happens when they
deviate. Jev's deviations select the truth 26.6\% of the time; the open models' deviations
select it 0.2--2.4\% of the time, close to noise. Re-tuning $\lambda$ does not rescue
them (8.09--8.37\%). Candidates are listed in the decoder's order, so position and
decoder confidence coincide at the top; the open models' answers track that position
rather than the sentences.

\subsection{Cost and latency}
\label{sec:cost}

\begin{table}[t]
\centering\small
\begin{tabular}{lrrrr}
\toprule
& \multicolumn{2}{c}{Cost (\$)} & \multicolumn{2}{c}{Latency per list (ms)} \\
\cmidrule(lr){2-3}\cmidrule(lr){4-5}
Arm & per 1k sentences & one user, per day & median & p95 \\
\midrule
Jev, API & 0.069 & 0.07 & 282 & 346 \\
OPT-6.7b, L40S & 0.029\rlap{$^*$} & 43.68 & 27 & 134 \\
Qwen2.5-7B, L40S & 0.034\rlap{$^*$} & 43.68 & 32 & 158 \\
\bottomrule
\end{tabular}
\caption{$^*$At full GPU utilisation. ``One user, per day'' assumes 1,000 sentences a
day and, for the 7B models, a dedicated on-demand GPU. Latency from the scoring runs, all
1,397 lists, one data centre in Finland; Jev's includes the network.}
\label{tab:cost}
\end{table}

Jev reads a mean 1,631 input tokens per list (95\% CI 1,398--1,868), which is \$0.069 per
thousand sentences (\cref{tab:cost}). A GPU running OPT-6.7b scores 62,000 lists per hour
when busy, \$0.029 per thousand, so a fully used GPU is cheaper per sentence. The GPU wins
only above 43\% utilisation, about 640,000 sentences a day. A single user speaks a small
fraction of that, so a dedicated GPU sits mostly idle and costs \$43.68 a day, while Jev
costs \$0.07 for a thousand sentences and needs no accelerator on the user's side.

Jev answers in a median 282\,ms against 27\,ms for OPT-6.7b on the local GPU. The London
measurement shows where the time goes. A full list takes a median 262\,ms end to end and
a two-option request 253\,ms, so list size barely matters. The provider's proxy reports a
median 62\,ms for both (p95 102\,ms; 77\,ms for lists of 60 or more candidates), leaving
about 194\,ms of transport between the client and the provider. The 62\,ms includes any
queueing inside the provider, so it bounds the model's own time from above. A co-located
deployment would therefore approach 60--100\,ms, the same order as OPT-6.7b's 27\,ms
median and 134\,ms p95, but not below it. We could not test this, because the model runs
only as a hosted service. At the roughly 32 words per minute reported for this
participant~\cite{card2024}, each sentence takes several seconds to attempt, so either
figure adds little to the wait.

\subsection{Limitations}
\label{sec:limits}

\begin{itemize}\itemsep2pt
\item \textbf{One closed model.} Jev's size and architecture are undisclosed, so the
  result shows that a typed-decision model can do this job, not why. We release its
  per-sentence label probabilities and settings so the analysis is reproducible, though
  the model is not. Hosted models change; our results are for \texttt{jev-latest} in
  September 2026.
\item \textbf{Post-hoc margin.} We chose the 0.2-point matching bound after seeing the
  results; a confirmatory study should fix it in advance.
\item \textbf{One participant, limited power.} T15 only, 978 test sentences; differences
  below about half a point are hard to resolve. 83 test sentences also occur, as text, in
  the phoneme network's training split, a property of the benchmark; excluding them
  changes no conclusion.
\item \textbf{Cloud dependence.} A hosted rescorer sends a user's decoded speech to a third
  party and fails without a network. A clinical device would need a local or private
  deployment, whose latency we could not measure.
\item \textbf{Prompt wording.} The typed prompt asks for the most fluent sentence, which
  could favour fluency over correctness. Jev picks the truth more often than the most
  fluent candidate when the two disagree, but we tested one wording.
\end{itemize}

\section{Conclusion}

On real intracortical recordings, one call to a typed-decision model rescores the
candidate list as well as a 7B language model that scores every candidate: Jev is ahead
of OPT-6.7b and Qwen2.5-7B under both tuning protocols and at most 0.2 points behind at
the 95\% bound. It costs cents per thousand sentences and removes the GPU from the user's
side; its latency over the internet is transport-bound and, at the provider, the same
order as a local 7B model.

Open general models given the same typed prompt fail, so the result depends on a model
trained for the decision. The next step is an open typed-decision model of known size
trained on this task, which would make both the cost and the latency claims testable on a
patient's own hardware.

\appendix

\section{Prompt}
\label{app:prompt}

Jev and Laya receive one request per list through the same interface, with candidates in
the decoder's order under labels A, B, and so on.

{\small
\begin{verbatim}
state = {"task": "A speech decoder produced these candidate sentences
                  for what a person was trying to say. Exactly one is
                  what they meant.",
         "candidates": {"A": <candidate 1>, "B": <candidate 2>, ...}}
questions = {"sentence": {
    "type": "choice",
    "instructions": "Which candidate is the sentence the person actually
                     meant? Judge by which reads as fluent, natural
                     English.",
    "criteria": {"A": <candidate 1>, "B": <candidate 2>, ...}}}
\end{verbatim}}

\noindent OPT-6.7b and Qwen2.5-7B receive the following text, wrapped in the model's chat
template where one exists and followed by \texttt{The answer is}; we read each label's
next-token probability.

{\small
\begin{verbatim}
A speech decoder produced these candidate sentences for what a person
was trying to say. Exactly one is what they meant. Answer with the
single label of the most fluent, natural English sentence.

A: <candidate 1>
B: <candidate 2>
...

Answer:
\end{verbatim}}

\end{document}